Large Language Models and Knowledge Graphs for Astronomical Entity Disambiguation

Golnaz Shapurian

Abstract

This paper presents an experiment conducted during a hackathon, focusing on using large language models (LLMs) and knowledge graph clustering to extract entities and relationships from astronomical text. The study demonstrates an approach to disambiguate entities that can appear in various contexts within the astronomical domain. By collecting excerpts around specific entities and leveraging the GPT-4 language model, relevant entities and relationships are extracted. The extracted information is then used to construct a knowledge graph, which is clustered using the Leiden algorithm. The resulting Leiden communities are utilized to identify the percentage of association of unknown excerpts to each community, thereby enabling disambiguation. The experiment showcases the potential of combining LLMs and knowledge graph clustering techniques for information extraction in astronomical research. The results highlight the effectiveness of the approach in identifying and disambiguating entities, as well as grouping them into meaningful clusters based on their relationships.

1. Introduction

Entity disambiguation is a critical challenge in processing astronomical text data, such as scientific publications, technical reports, and observational logs. For instance, the entity "Apollo" can refer to missions, craters, or even an asteroid. Accurate disambiguation of such entities is essential for efficient information retrieval, knowledge discovery, and text mining within the astronomical domain. Failure to do so can lead to confusion, misinterpretation of information, and hinder research progress (Brescia and Longo, 2012).

Knowledge graphs (KGs) have emerged as a powerful tool for addressing this challenge by representing and organizing complex information in a structured and machine-readable format. In astronomical research, KGs can integrate data from diverse sources, uncover hidden patterns and relationships, and support decision-making (Khoo et al., 2023). By representing entities and their relationships in a graph structure, researchers can gain new insights, formulate hypotheses, and promote knowledge sharing across different subdomains of astronomy.

Large language models (LLMs), such as GPT-4, trained on extensive textual datasets, exhibit exceptional proficiency in comprehending and producing human-like language. By harnessing the knowledge captured in LLMs, it is possible to extract entities and relationships from text with high accuracy and efficiency, reducing the reliance on manual annotation and rule-based approaches.

This experiment investigates the use of the GPT-4 language model to extract entities and relationships from astronomical text, focusing on entity disambiguation. A knowledge graph is constructed from the extracted information to capture the connections between various astronomical concepts. The Leiden algorithm is then applied to cluster the knowledge graph, identifying communities of related entities and relationships (Traag et al., 2019; Edge et al., 2024). These communities are used to disambiguate unknown excerpts by calculating their association percentages.

By evaluating the effectiveness of this approach, this study demonstrates the potential of combining LLMs and knowledge graph clustering techniques for entity disambiguation and information extraction in astronomical text. The findings can contribute to developing more efficient and accurate tools for knowledge discovery and exploration in astronomy, leading to better information retrieval, more precise data integration, and a deeper understanding of astronomical phenomena. This approach streamlines the process of extracting insights from astronomical literature, ultimately accelerating scientific progress.

2. Related Work

The integration of LLMs and KGs has gained significant attention as researchers explore combining these two powerful technologies. LLMs, such as GPT-4, have demonstrated remarkable abilities in natural language processing and generation, while KGs provide structured representations of knowledge that can enhance reasoning and inference capabilities.

Pan et al. (2023) present a roadmap for the unification of LLMs and KGs, outlining three general frameworks: KG-enhanced LLMs, LLM-augmented KGs, and Synergized LLMs + KGs. They review existing efforts within these frameworks and highlight future research directions, emphasizing the complementary nature of LLMs and KGs. Shu et al. (2024) introduce the Knowledge Graph Large Language Model Framework, leveraging chain-of-thought prompting and in-context learning to enhance multi-hop link prediction in KGs.

Several studies have explored LLMs in supporting Knowledge Graph Engineering (KGE). Meyer et al. (2023) demonstrate how ChatGPT can assist in the development and management of KGs. Trajanoska et al. (2023) analyze how advanced LLMs, like ChatGPT, compare with specialized pretrained models for joint entity and relation extraction. They find that using advanced LLM models can improve the accuracy of the automatic creation of KGs from raw texts.

Jia et al. (2024) propose a framework integrating the Deep Document Model and KG-enhanced Query Processing for optimized complex query handling in scholarly knowledge graphs. Their approach, combining the ANU Scholarly Knowledge Graph with LLMs, enhances knowledge utilization and natural language understanding.

The use of KGs as factual context for LLM prompts has also been investigated. Abu-Rasheed et al. (2024) explore this concept in personalized education, utilizing semantic relations in the KG to offer curated knowledge about learning recommendations. Their results show enhanced recall and precision of generated explanations compared to those generated solely by GPT.

Yang et al. (2023) review studies on integrating pre-trained language models with KGs, detailing existing KG-enhanced models and applications. They propose developing KG-infused LLMs to further improve their factual reasoning ability.

In astronomical research, Sun et al. (2024) demonstrate using LLMs to extract concepts from publications and form a KG. Their knowledge graph reveals two phases of development: technology integration and exploration in scientific discovery.

Building upon Trajanoska et al. (2023), this experiment extends their approach by incorporating entity disambiguation and contextual understanding within the astronomical domain. While their study focused on KG construction, this work leverages LLMs' natural language processing strengths and KGs' structured knowledge representation to enhance the accuracy and relevance of knowledge extraction from astronomical text. The key contribution lies in applying the Leiden algorithm for community detection within the KG, enabling effective disambiguation of entities in different astronomical contexts. By integrating advanced clustering techniques with KG construction, this study aims to improve the precision and contextual understanding of extracted entities, ultimately facilitating more accurate and efficient knowledge discovery in astronomy.

3. Data Collection

The data collection process involved querying the Astrophysics Data System (ADS) using its application programming interface (Kurtz et al., 2000) to gather textual data related to the astronomical entity "Apollo." Two separate queries were performed to capture different contexts in which the term appears.

The first query focused on retrieving documents related to Apollo craters on the Moon by searching for the full text of articles containing the words "Apollo," "Crater," and "Moon," limited to planetary science journals.

The second query aimed to gather documents related to Apollo missions, searching for the full text of articles containing "Apollo," "mission," and any Apollo mission name from a Wikipedia-sourced list.

The records from both queries were combined, and a regular expression-based approach extracted excerpts of at most 256 words, consisting of complete sentences, surrounding each occurrence of "Apollo." Quality control measures, including manual review to filter out irrelevant or low-quality documents and maintain balance between Apollo crater and Apollo mission contexts, resulted in approximately 800 excerpts.

No extensive data cleaning or preprocessing was performed on the extracted excerpts to preserve the authenticity and contextual information associated with each occurrence of the term "Apollo."

This curated collection of excerpts, each containing the term "Apollo" and its surrounding context from both Apollo crater and Apollo mission-related documents, served as the foundation for subsequent relationship extraction and entity disambiguation using large language models and knowledge graph clustering techniques.

4. Methodology

The experiment methodology involved using the GPT-4 language model for entity and relationship extraction, constructing a knowledge graph, and applying the Leiden algorithm for community detection and entity disambiguation.

The GPT-4 language model was accessed through the OpenAI API, using a specialized prompt containing instructions and two examples relevant to astronomy (see Appendix). The prompt, adapted from Trajanoska et al. (2023), included representative examples showcasing the desired input-output format and types of entities and relationships to be extracted. One example focused on Apollo missions, while the other revolved around Apollo craters. Each example consisted of a short text passage, followed by a structured representation of the extracted entities and their relationships, labeled with predefined categories (e.g., celestial objects, missions, instruments, spacecraft) and their semantic connections.

To ensure focused and deterministic responses, the GPT-4 model's temperature parameter was set to 0. The 256-word excerpts containing the term "Apollo" were then processed by the model, extracting relevant entities and relationships based on the astronomical context guided by the prompt.

A knowledge graph was constructed to organize the extracted entities and relationships, with nodes representing entities and edges representing relationships. This facilitated visualization and exploration of the interconnections between different astronomical concepts related to Apollo craters and Apollo missions.

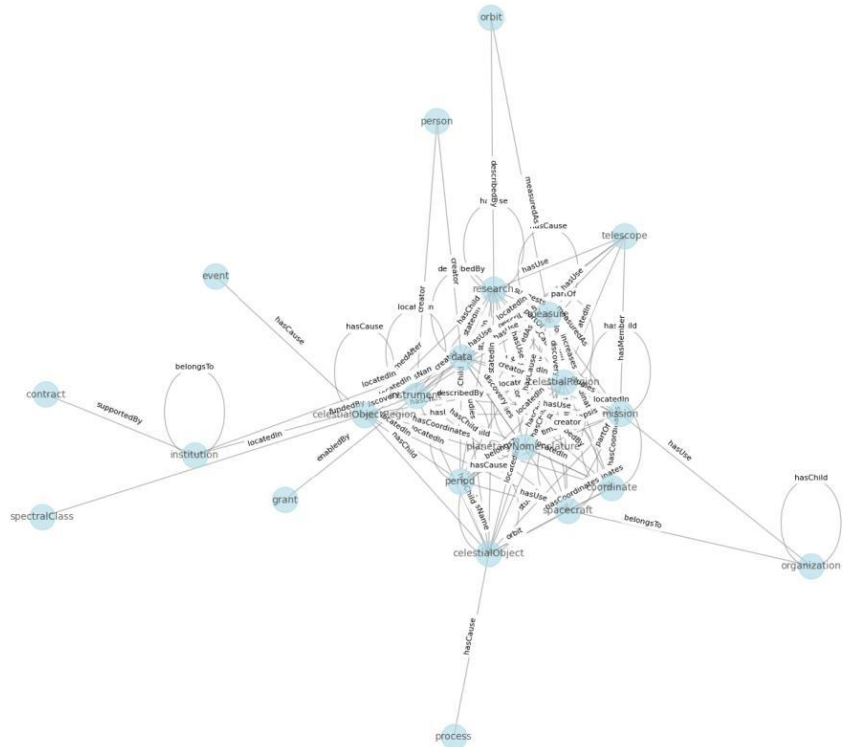

Figure 1: Knowledge graph constructed from the Apollo dataset, showing relationships between entities such as celestial objects, missions, and instruments, with nodes representing entities and edges representing their relationships.

The Leiden algorithm was applied for community detection within the knowledge graph. This clustering technique identified groups of densely connected nodes, uncovering clusters of semantically related entities and relationships likely belonging to the same context (e.g., Apollo craters or Apollo missions).

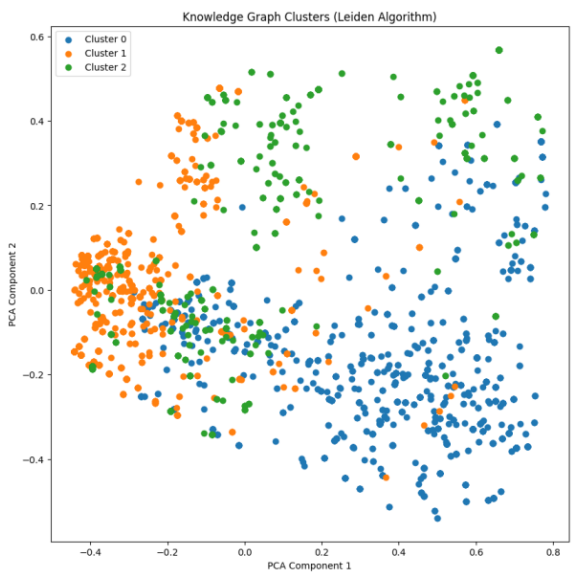

Figure 2: Clusters identified by the Leiden algorithm within the knowledge graph, highlighting three distinct groups of densely connected entities related to Apollo missions, and craters.

Entity disambiguation was performed using the detected communities. When a new excerpt containing the term "Apollo" was encountered, its extracted entities and relationships were compared against existing communities. By calculating the percentage of association between the new excerpt and each community, the most likely context for the term "Apollo" was determined, effectively disambiguating whether it referred to a crater or a mission.

5. Results and Discussion

The experiment results demonstrate the effectiveness of combining large language models and knowledge graph clustering for entity disambiguation and contextual understanding in astronomy. The clustering algorithm successfully identified three distinct clusters based on entity associations with Apollo craters and missions.

Table 1 presents the distribution of crater and mission entities within each cluster. Cluster 0 was predominantly associated with mission entities (73.63%), while Clusters 1 and 2 were more strongly associated with crater entities (92.46% and 81.33%, respectively). The high percentages of crater-related entities in Clusters 1 and 2 indicate the approach's success in disambiguating and grouping entities from crater excerpts.

| Cluster | Crater Entities (%) | Mission Entities (%) |
|---|---|---|
| 0 | 26.37% | 73.63% |
| 1 | 92.46% | 7.54% |
| 2 | 81.33% | 18.67% |

Table 1: Distribution of crater and mission entities within each cluster identified by the clustering algorithm, showing that Cluster 0 is predominantly associated with mission entities, while Clusters 1 and 2 are more strongly associated with crater entities.

To evaluate disambiguation performance, a confusion matrix comparing the predicted and true labels of test entities was constructed and is displayed in Table 2.

|  | Predicted Crater | Predicted Mission |
|---|---|---|
| Actual Crater | 1159 | 293 |
| Actual Mission | 391 | 418 |

Table 2: Confusion matrix evaluating the performance of the entity disambiguation approach, showing the predicted versus actual labels for crater and mission entities.

Several evaluation metrics were calculated. The overall accuracy was 69.76%. For the crater class, precision was 74.77%, recall was 79.82%, and F1 score was 77.21%. For the mission class, precision was 58.78%, recall was 51.67%, and F1 score was 55.05%. This indicates better performance in predicting crater instances compared to mission instances.

The results highlight the effectiveness of the approach in disambiguating entities and understanding their contextual relationships in astronomy. The clustering algorithm successfully grouped entities based on their associations with Apollo craters and missions, providing a clear separation between the two contexts. The evaluation metrics further validate the model's performance, particularly in identifying and classifying crater-related entities. However, the lower performance for mission-related entities suggests potential for improvement in capturing their complexities.

6. Conclusion

This study demonstrates an approach to entity disambiguation and contextual understanding in the astronomical domain by leveraging large language models and knowledge graph clustering techniques. The combination of the GPT-4 language model for entity and relationship extraction with the Leiden algorithm for community detection successfully disambiguates entities related to Apollo craters and Apollo missions in astronomical text data.

The experimental results highlight the effectiveness of the approach, with the clustering algorithm identifying distinct clusters associated with crater and mission entities. Evaluation metrics demonstrate the model's commendable performance in disambiguating and classifying crater-related entities, while acknowledging the potential for improvement in capturing the complexities of mission-related entities.

This work contributes to the growing body of research on applying advanced natural language processing and knowledge graph-based approaches in astronomy. The proposed methodology offers a promising direction for improving information retrieval, data integration, and knowledge organization in astronomical text data, ultimately leading to deeper insights and accelerated scientific progress.

Appendix

The following prompt was used to fine-tune the GPT-4o language model for entity and relationship extraction in the context of Apollo craters and Apollo missions. The prompt consists of two examples, each containing a short passage of text and a structured representation of the extracted entities and their relationships. These examples were carefully selected to showcase the desired input-output format and the types of entities and relationships relevant to the astronomical domain.

```
"""
Based on the following two examples, extract entities and relations from the provided
text.
Use the following entity types:
# ENTITY TYPES:
{
"instrument": "https://www.wikidata.org/wiki/Q751997", # astronomical instrument
"telescope": "https://www.wikidata.org/wiki/Q148578", # space telescope
"mission": "https://www.wikidata.org/wiki/Q2133344", # space mission
"celestialObject": "https://www.wikidata.org/wiki/Q6999", # astronomical object
"celestialRegion": "https://www.wikidata.org/wiki/Q203218", # spherical coordinate
system
"celestialObjectRegion": "https://www.wikidata.org/wiki/Q12134", # celestial sphere
"planetaryNomenclature": "https://www.wikidata.org/wiki/Q1463003", # planetary
nomenclature: system of uniquely identifying features on the surface of a planet or
natural satellite
"spacecraft": "https://www.wikidata.org/wiki/Q40218", # spacecraft
"measure": "https://www.wikidata.org/wiki/Q39875001", # measure: standard against which
something can be judged
"period": "https://www.wikidata.org/wiki/Q392928", # period: subdivision of geological
time; shorter than an era and longer than an epoch
"coordinate": "https://www.wikidata.org/wiki/Q3250736", # coordinate: number which
characterizes position
"research": "https://www.wikidata.org/wiki/Q42240", # research: systematic study
undertaken to increase knowledge
"data": "https://www.wikidata.org/wiki/Q42848", # data: information arranged for
automatic processing
}

Use the following relation types:
{
"discovery": "https://www.wikidata.org/wiki/Q12772819", # discovery
"studies": "http://www.wikidata.org/prop/direct/P2579", # studied in
"measuredAs": "http://www.wikidata.org/prop/direct/P111", # measured physical
quantity
"hasChild": "http://www.wikidata.org/prop/direct/P40", # child
"describedBy": "http://www.wikidata.org/prop/direct/P1343", # described by source
"locatedIn": "http://www.wikidata.org/prop/direct/P276", # location
"hasUse": "http://www.wikidata.org/prop/direct/P366", # use
"orbitalInclination": "http://www.wikidata.org/prop/direct/P2045", # orbital
inclination
"hasCoordinates": "http://www.wikidata.org/prop/direct/P625", # coordinate location
"statedIn": "https://www.wikidata.org/wiki/Property:P248", # to be used in the
references field to refer to the information document or database in which a claim is
made
"creator": "https://www.wikidata.org/wiki/Property:P170", # creator
"hasCause": "https://www.wikidata.org/wiki/Property:P828", # corrolates
}
```

–> Beginning of example 1

# Specification
From both the Apollo and Surveyor experience, it is known that small craters, rather than blocks, may actually present a greater danger to landing vehicles because they are far more abundant. In fact, except for very young Copernican units, the lunar surface is saturated with craters smaller than 100 m in diameter (Gault 1970). Thus, a lunar landing site will contain many craters, presenting a sub-horizontal surface at the scale of any lander. Blocks, on the other hand, are much less common, especially in areas away from fresh craters (Cameron and Coyle 1971; Cintala et al. 1982; Cintala and McBride 1995; Lee et al. 1986; Moore 1971). The presence of small, shallow craters can be assessed from images with large ($> 70°$) incidence angles and meterscale topography.

#################

# Output
[
{{
"head": "Apollo",
"head_type": "spacecraft",
"relation": "discovery",
"tail": "small craters present a greater danger to landing vehicles",
"tail_type": "celestialObjectRegion"
}},
{{
"head": "Surveyor",
"head_type": "spacecraft",
"relation": "discovery",
"tail": "small craters present a greater danger to landing vehicles",
"tail_type": "celestialObjectRegion"
}},
{{
"head": "Apollo",
"head_type": "spacecraft",
"relation": "studies",
"tail": "Moon",
"tail_type": "celestialObject"
}},
{{
"head": "Surveyor",
"head_type": "spacecraft",
"relation": "studies",
"tail": "Moon",
"tail_type": "celestialObject"
}},
{{
"head": "Copernican units",
"head_type": "period",
"relation": "measuredAs",
"tail": "lunar surface",
"tail_type": "celestialObjectRegion"
}},
{{
"head": "Moon",
"head_type": "celestialObject",
"relation": "hasChild",
"tail": "lunar poles",
"tail_type": "celestialObjectRegion"
}},
{{
"head": "Moon",
"head_type": "celestialObject",

```
"relation": "hasChild",
"tail": "lunar surface",
"tail_type": "celestialObjectRegion"
}},
{{
"head": "lunar surface",
"head_type": "celestialObjectRegion",
"relation": "describedBy",
"tail": "craters smaller than 100 m in diameter",
"tail_type": "celestialObject"
}},
{{
"head": "lunar landing site",
"head_type": "celestialObjectRegion",
"relation": "describedBy",
"tail": "many craters, presenting a sub-horizontal surface at the scale of any lander",
"tail_type": "celestialRegion"
}},
{{
"head": "blocks",
"head_type": "celestialRegion",
"relation": "locatedIn",
"tail": "areas away from fresh craters",
"tail_type": "celestialRegion"
}},
]
```

–> End of example 1
–> Beginning of example 2

# Specification
Using this Apollo-based "calibration" of crater degradation, Basilevsky et al. (2013) employed high-resolution images from the Lunar Reconnaissance Orbiter Camera (LROC) to analyze the frequency of meter-scale boulders in the ejecta deposits of these dated Apollo craters, as well as from a few other craters that were of intermediate or more advanced degradational state and relative age based on Basilevsky (1976). The number density (number per unit surface area) of boulders >2 m observed in the ejecta deposits of these craters correlates well with absolute crater-formation age(s) as illustrated in Fig. 9.

################

# Output
```
[
{{
"head": "Basilevsky et al. (2013)",
"head_type": "research",
"relation": "statedIn",
"tail": "high-resolution images from LROC",
"tail_type": "data"
}},
{{
"head": "high-resolution images",
"head_type": "data",
"relation": "creator",
"tail": "Lunar Reconnaissance Orbiter Camera (LROC)",
"tail_type": "instrument"
}},
{{
"head": "Apollo craters",
"head_type": "planetaryNomenclature",
"relation": "hasChild",
```

```
"tail": "ejecta deposits",
"tail_type": "celestialRegion"
}},
{{
"head": "number density of boulders >2 m",
"head_type": "measure",
"relation": "locatedIn",
"tail": "ejecta deposits of craters",
"tail_type": "celestialRegion"
}},
{{
"head": "number density of boulders >2 m",
"head_type": "measure",
"relation": "hasCause",
"tail": "absolute crater-formation age(s)",
"tail_type": "measure"
}},
{{
"head": "meter-scale boulders",
"head_type": "celestialObject",
"relation": "locatedIn",
"tail": "ejecta deposits",
"tail_type": "celestialRegion"
}},
{{
"head": "ejecta deposits",
"head_type": "celestialRegion",
"relation": "hasChild",
"tail": "Apollo craters",
"tail_type": "planetaryNomenclature"
}},
{{
"head": "ejecta deposits",
"head_type": "celestialRegion",
"relation": "hasChild",
"tail": "craters of intermediate or more advanced degradational state and relative age",
"tail_type": "celestialObject"
}},
]
-> End of example 2

For the following specification, generate and extract entities and relations as in the provided example.

# Specification
{specification}
################

# Output
"""
```

The prompt was designed to provide the GPT-4o model with representative examples that demonstrate the specific terminology, writing styles, and context of astronomical literature. By training the model on these examples, it learned to recognize the patterns and structures necessary for accurate entity and relationship extraction in the given context.